\documentclass{sig-alternate-05-2015}

\usepackage{float}
\usepackage{times}
\usepackage{fixltx2e}
\usepackage{algorithm,algorithmic}
\usepackage{booktabs}

\newcommand{\defeq}{\stackrel{\text{def}}{=}}

\renewcommand{\(}{\left(}
\renewcommand{\)}{\right)}

\newcommand{\Hk}[1]{\mathcal{H}_K \( #1 \)}
\newcommand{\sgn}[1]{\mathrm{sgn}\( #1 \)}

\newcommand{\st}{\mathrm{s.t.}}

\newcommand{\ba}{\boldsymbol{a}}
\newcommand{\R}{\mathbb{R}}
\newcommand{\Rd}{\mathbb{R}^d}

\newcommand{\twonorm}[1]{\left\lVert #1 \right\rVert_{2}}
\newcommand{\onenorm}[1]{\left\lVert #1 \right\rVert_{1}}
\newcommand{\zeronorm}[1]{\left\lVert #1 \right\rVert_{0}}
\newcommand{\abs}[1]{\left\lvert #1 \right\rvert}

\begin{document}

\title{Learning Large Scale Sparse Models}

%
%
%
%
%

\numberofauthors{3} 
%
\author{
%
%
\alignauthor
Atul Dhingra\\
    \email{atul.dhingra@rutgers.edu}
\alignauthor    
Jie Shen\\
    \email{js2007@rutgers.edu}
\alignauthor 
Nicholas Kleene\\
    \email{nk374@rutgers.edu}
}

\maketitle

%
%

%
%

\section*{Abstract}
In this work, we consider learning sparse models in large scale settings, where the number of samples and the feature dimension can grow as large as millions or billions. Two immediate issues occur under such challenging scenario: (i) computational cost; (ii) memory overhead. In particular, the memory issue precludes a large volume of prior algorithms that are based on batch optimization technique. To remedy the problem, we propose to learn sparse models such as Lasso in an online manner where in each iteration, only one randomly chosen sample is revealed to update a sparse iterate. Thereby, the memory cost is independent of the sample size and gradient evaluation for one sample is efficient. Perhaps amazingly, we find that with the same parameter, sparsity promoted by batch methods is not preserved in online fashion. We analyze such interesting phenomenon and illustrate some effective variants including mini-batch methods and a hard thresholding based stochastic gradient algorithm. Extensive experiments are carried out on a public dataset which supports our findings and algorithms.
\section*{Keywords} Sparsity, Lasso, Large Scale Algorithm, Online Optimization
\section{Introduction}
In the face of big data, learning structured representation continues to be one of the most quintessential tasks in machine learning. In this paper, we are interested in such large scale setting, i.e., we have to manipulate thousands or millions of features and perhaps billions of samples.

To tackle the curse of dimensionality, Lasso was proposed for variable selection~\cite{tibshirani1996regression}, where the $\ell_1$ constraint was utilized since it is a tight convex approximation to the $\ell_0$ norm.

On the other hand, in order to learn models based on huge scale datasets, one promising alternative is stochastic optimization techniques. Compared to batch algorithms that require to access all the samples during optimization, stochastic methods such as stochastic gradient descent~(SGD)~\cite{zhang2004sgd}, stochastic average gradient~(SAG)~\cite{sag} and stochastic variance reduced gradient~(SVRG)~\cite{svrg} randomly pick one sample in each iteration for updating the parameters, hence the memory cost is independent of sample size. Interestingly, for the high dimensional case where the number of features is comparable or even larger than the dataset size, some dual coordinate ascent algorithms have been proposed to mitigate the computation, such as stochastic coordinate dual ascent~(SCDA)~\cite{shalev2013stochastic} and its accelerated version~\cite{shalev2014accelerated}. Recently, it becomes clear that all the methods mentioned can be formulated in the variance reduction framework~\cite{saga}, which facilitates a unified analysis and new hybrid algorithms~\cite{reddi2015variance}.

In order to simultaneously address both issues~(high dimensionality and huge sample size), it is natural to devise online algorithms for Lasso. Some prominent examples include forward-backward splitting~(Fobos)~\cite{fobos}, regularized dual averaging~(RDA) \cite{xiao2009dual} and proximal algorithms~\cite{saga,xiao2014proximal}. However, {\bf none of them is capable of producing sparse solutions in an online setting, even though they are optimizing the $\ell_1$ norm}. The intuition why Lasso fails in online manner is that, in each iteration, we perform a gradient update followed by soft-thresholding~(i.e. the Lasso step). But, note that since we only pick one sample to compute the gradient, the magnitude of the gradient is small, and hence a large $\lambda$ would shrink all of them to zero, while a small $\lambda$ cannot promote sparsity! Since each sample is chosen randomly, it is essentially hard to find a proper penalty $\lambda$. In contrast, batch methods will not incur such issues since the full dataset is always loaded to compute the gradient. The coordinate descent algorithm was recently shown to achieve linear convergence for strongly convex and composite objectives, and sublinear when only convexity is guaranteed~\cite{richtarik2014iteration}. While promising, this approach is not practical for coping with huge sample size.

\subsection{Related Works}
\vspace{0.05in}
\noindent{\bf Dictionary Learning.} \ 
There is a large body of works devoted to learning sparse models under different settings. For example, in machine learning and computer vision, dictionary learning~(also known as sparse coding) aims to simultaneously optimize an over-complete dictionary and the corresponding codes for a given dataset~\cite{wright2010sparse,mairal2010online}. There, the sparsity pattern naturally emerges since the dictionary is over-complete. Recently, a theoretical study on the sample complexity and local convergence was carried out which partially explains the success of the empirical algorithm~\cite{vainsencher2011sample,agarwal2013learning,arora2013new}.

\vspace{0.05in}
\noindent{\bf Compressed Sensing.} \ 
Parallel to the community of dictionary learning, the core problem considered by compressed sensing is inverting the linear data acquisition process by using many fewer measurements than the signal capacity, i.e., solving a linear system where the unknowns are much larger than the equations~\cite{donoho2006compressed}. In general, there are infinite number of solutions fitting such system and we have no hope to recover the true signal. Yet, knowing that the signal of interest is sparse immediately changes the premise. In particular, \cite{candes2005decoding} showed that when the sensing matrix satisfies the restricted isometry property~(RIP), then one may exactly recover the underlying signal by solving an $\ell_1$ based program. This is, perhaps, the first known result that the convex relaxed program is equivalent to the NP hard problem. Based on the RIP condition, numerous practical algorithms were developed to establish exact sparse signal recovery, for example, orthogonal matching pursuit~(OMP)~\cite{tropp2007signal}, iterative hard thresholding~(IHT)~\cite{blumensath2009iterative}, compressive sampling matching pursuit~(CoSaMP)~\cite{needell2009cosamp}, subspace pursuit~(SP)~\cite{dai2009subspace}, to name just a few. One may refer to a comprehensive review~\cite{tropp2010computational} and the reference therein for more details.

Underlying most of these algorithms is a simple yet effective operator termed hard thresholding, which sets all but the largest $k$ absolute values to zero. Hence, it is a good fit to solve the $\ell_0$ constrained program. \cite{yuan2014gradient,bahmani2013greedy} extends this technique to the machine learning setting, where the objective function is a general loss. Yet, they are all based on a batch setting, i.e. their algorithms and theoretical results only hold if all samples are revealed in each iteration. It still remains an open question if a $\ell_0$ constrained formulation can be exactly solved in the online setting. The challenge falls into the fact that online optimization introduces variance to the gradient evaluation which is hard to characterize.

\vspace{0.05in}
\noindent{\bf Beyond.} Examples of fields that explore sparsity are exhaustive. In computer vision, researchers pursue sparsity for dictionary learning as we aforementioned, and sparse subspace clustering~\cite{elhamifar2009sparse}, sparse representation~\cite{wright2009robust}, uniform or column-wise sparse corruption~\cite{candes2011robust,xu2010robust}, along with numerous algorithms such as~\cite{lee2006efficient}. In the regime of missing entries, interesting works include sparse graph clustering~\cite{chen2012clustering}, partially observed graphs~\cite{chen2014clustering} and matrix completion~\cite{candes2009exact,candes2010power}.

\subsection{Summary of Contributions}
We propose to tackle the important but challenging problem of producing sparse solutions for large scale machine learning. Compared to the batch methods, our formulation reduces the memory cost by one order of magnitude. Compared to existing methods, our empirical study demonstrates that our improved Lasso formulation and hard thresholded SGD algorithm will always result in sparse solutions.

\section{Algorithms}
In this work, we are mainly interested in the Lasso formulation which has been known to promote sparse solutions. In particular, given a set of training samples $\{ (\ba_i, y_i) \}_{i=1}^n \subset \Rd \times \R$, Lasso pursues a sparse solution by optimizing the following program:
\begin{align}\label{eq:primal}
\min_{x \in \Rd}\quad F(x) \defeq \frac{1}{2n} \twonorm{Y - Ax}^2 + \lambda \onenorm{x},
\end{align}
where $Y = (y_1, y_2, \dots, y_n)^{\top}$, $A = (\ba_1; \ba_2; \dots; \ba_n)$ and $\lambda$ is a tunable positive parameter. We also note that the objective $F(x)$ is decomposable, i.e.,
\begin{align}
F(x) = \frac{1}{n} \sum_{i=1}^n f_i(x),
\end{align}
where
\begin{align}
f_i(x) \defeq \frac{1}{2} \twonorm{y_i - \ba_i \cdot x}^2 + \lambda \onenorm{x}.
\end{align}

\subsection{Coordinate Descent}
Coordinate descent~(CD) is one of the most popular methods for solving the Lasso problem. In each iteration, it updates the coordinates of $x$ in a sequential (or random) manner by fixing the remaining coordinates. Let $x^{t-1}$ be the iterate at the $t$-th iteration. Denote
\begin{align}\label{eq:M_i}
M_i = Y - \sum_{j \neq i} A_j x_j^{t-1},\quad 1 \leq i \leq d.
\end{align}
where we write the $j$th column of $A$ as $A_j$. In this way, solving the $i$th coordinate of $x^t$ amounts to minimizing
\begin{align}
g(x) = \frac{1}{2n} \twonorm{M_i - A_i x}^2 + \lambda \abs{x},
\end{align}
where the closed-form solution is given by
\begin{align}\label{eq:x_i}
x_i^t = \twonorm{A_i}^{-2} \mathcal{S}_{n\lambda}(A_i^{\top} M_i), \quad 1 \leq i \leq d.
\end{align}
Here, $\mathcal{S}_{\lambda}(v)$ is the soft-thresholding operator:
\begin{align}
\mathcal{S}_{\lambda}(v) = \begin{cases}
v - \lambda, \quad &\textrm{if}\ v > \lambda,\\
v + \lambda,\quad &\textrm{if}\ v < -\lambda,\\
0,\quad &\textrm{otherwise}.
\end{cases}
\end{align}

We summarize the coordinate descent algorithm in Algorithm~\ref{alg:cd}.
\begin{algorithm}[H]
\caption{Coordinate Descent}
\begin{algorithmic}[1]\label{alg:cd}
\REQUIRE Training sample $(A, Y)$, parameter $\lambda$, maximum iteration count $T$.
\ENSURE Optimal solution $x^*$ to~\eqref{eq:primal}.
\STATE Initialize the iterate $x^0$ with zero vector.
\FORALL{$t=1, 2, \dots, T$}
\FORALL{$i=1, 2, \dots, d$}
\STATE $M_i = Y - \sum_{j \neq i} A_j x_j^{t-1}$.
\STATE $x_i^t = \twonorm{A_i}^{-2} \mathcal{S}_{n\lambda}(A_i^{\top} M_i)$.
\ENDFOR
\ENDFOR
\STATE Set $x^* = x^T$.
\end{algorithmic}
\end{algorithm}

\vspace{0.05in}
\noindent{\bf Positive.} The CD algorithm is easy to implement, and usually produces sparse solutions for proper $\lambda$. Its convergence behavior was recently carried out which illustrates that a stochastic version of CD enjoys linear convergence if strong convexity of the objective function is satisfied and a sublinear convergence is attained for general convex functions~\cite{richtarik2014iteration}

\vspace{0.05in}
\noindent{\bf Negative.} In order to update the coordinate, the algorithm has to load all the samples along that dimension, for which the memory cost is $O(n)$. This is not practical when we are dealing with huge scale datasets.

\subsection{Sub-Gradient Descent}
Parallel to coordinate descent, gradient descent is another prominent algorithm in mathematical optimization~\cite{bertsekas1999nonlinear,nesterov2013introductory}. That is, for a differentiable convex function $h(x)$, one computes its gradient evaluated at $x = x^{\textrm{old}}$ and update the solution as
\begin{align}
x^{\textrm{new}} = x^{\textrm{old}} - \eta \cdot \nabla h(x^{\textrm{old}})
\end{align}
for some pre-defined learning rate $\eta > 0$. However, the $L_1$ norm is not differentiable at $x = 0$ so we have resort to the sub-gradient method. While for first-order differentiable functions the gradient is uniquely determined at a given point, the sub-gradients of a non-differentiable convex function $h(x)$ at $x = x'$ form a set as follows:
\begin{align}
\frac{\partial}{\partial x} h(x)\vert_{x = x'} = \{ z \mid h(x) - h(x') - z^{\top}(x - x') \geq 0 \}.
\end{align}

In our problem, the $L_1$ norm $\onenorm{x}$ is non-differentiable and one may verify that its sub-gradients at $x = 0$ is given by
\begin{align}
\frac{\partial}{\partial x} \onenorm{x}\vert_{x = 0} = \{ z \mid -1 \leq z_i \leq 1, 1 \leq i \leq d \}.
\end{align}
In our implementation, we simply choose the zero vector as the sub-gradient if $x = 0$. Hence, if not specified, we use the sub-gradient
\begin{align}
z = \sgn{x}
\end{align}
for the $L_1$ norm.

\vspace{0.05in}
\noindent{\bf Batch Method.} The batch version of sub-gradient descent for~\eqref{eq:primal} is easy to implement. Given an old iterate $x^{t-1}$, we compute the sub-gradient of the objective function as follows:
\begin{align}
z^t = \frac{1}{n} A^{\top}(Ax^{t-1} - Y) + \lambda \cdot \sgn{x^{t-1}}.
\end{align}
Then we make gradient update
\begin{align}
x^t = x^{t-1} - \eta z^t.
\end{align}
Notably, we can accelerate the computation of full gradient by storing the matrix $A^{\top}A$ and the vector $A^{\top}Y$, which is shown in Algorithm~\ref{alg:sbgd-batch}.
\begin{algorithm}[H]
\caption{Full Sub-Gradient Descent}
\begin{algorithmic}[1]\label{alg:sbgd-batch}
\REQUIRE Training sample $(A, Y)$, parameter $\lambda$, maximum iteration count $T$.
\ENSURE Optimal solution $x^*$ to~\eqref{eq:primal}.
\STATE Compute $B = A^{\top}A$ and $C = A^{\top} Y$.
\STATE Initialize $F^* = F(0)$.
\FORALL{$t=1, 2, \dots, T$} 
\STATE $x^{t}=x^{t-1}- \eta \big[n^{-1} (Bx^{t-1}-C) + \lambda \cdot \sgn{x^{t-1}}\big]$.
\IF{$F(x^t) < F^*$}
\STATE $x^* \leftarrow x^t,\quad F^* \leftarrow F(x^t)$.
\ENDIF
\ENDFOR
\end{algorithmic}
\end{algorithm}

\vspace{0.05in}
\noindent{\bf Positive.} Again, the sub-gradient method is easy to follow and a sublinear convergence rate, i.e., $O(1/T)$ was established for our non-smooth objective function~\cite{nesterov2004introductory}.

\vspace{0.05in}
\noindent{\bf Negative.} The memory cost is $O(nd)$ that is not practical for large scale datasets. Also note that our goal is to learning a sparse solution $x^*$. During the optimization, sub-gradient descent method {\em always} computes the sub-gradient and adds it to the iterate scaled by the step size. Hence, there is no shrinking operation to make the solution sparse. Recall that CD invokes soft-thresholding each time. Thereby, chances are that the solution produced by sub-gradient methods are not exactly sparse, i.e., very small values such as $10^{-10}$ may be preserved.

\vspace{0.05in}
\noindent{\bf Mini-Batch and Stochastic Variant.} Due to the memory issue, we also consider the mini-batch and stochastic variant of sub-gradient descent. Note that the stochastic variant is virtually a special case of mini-batch when the batch size is one. So we instate the mini-batch implementation.

To begin, let $\Lambda^t = \{ i^t_1, i^t_2, \dots, i^t_m \} \subset \{1, 2, \dots, n\}$ be an index set which is drawn randomly at the $t$-th iteration with batch size $m$. Let $A_{\Lambda^t}$ be the submatrix of $A$ consisting of the rows of $A$ indexed by $\Lambda^t$. Likewise, we define $Y_{\Lambda^t}$. With the notation on hand, the mini-batch sub-gradient method can be stated as follows:
\begin{align}
z^t &= \frac{1}{m} A^{\top}_{\Lambda^t} \( A_{\Lambda^t} x^{t-1} - Y_{\Lambda^t} \) + \lambda \cdot \sgn{x^{t-1}},\\
x^t &= x^{t-1} - \eta_t \cdot z^t.
\end{align}

\vspace{0.05in}
\noindent{\bf Remark.} \ 
Note that there are two differences between the full sub-gradient method and the mini-batch (or stochastic) one. First, the gradient of the latter one has randomness so the convergence is guaranteed in expectation. Second, due to the randomness, or more precisely, the variance, the step size $\eta_t$ has to decay to zero with $t$, while we can choose a constant step size $\eta$ for the batch version. The rationale is that, when the sequence $\{x^t\}$ converges, $\eta_t \cdot z^t$ tends to zero. Since $z^t$ is not the full gradient, it cannot be zero. Hence, $\eta_t$ has to converge to $0$. Interestingly, recent works are devoted to designing variance reduction algorithms which allow the learning rate to be a constant, see, e.g.,~\cite{svrg}. In our work, we pick $\eta_t = 1/t$.

For the reader's convenience, we write the algorithm in Algorithm~\ref{alg:sbgd-mini}.
\begin{algorithm}[H]
\caption{Mini-Batch/Stochastic Sub-Gradient Descent}
\begin{algorithmic}[1]\label{alg:sbgd-mini}
\REQUIRE Training sample $(A, Y)$, parameter $\lambda$, maximum iteration count $T$, batch size $m$.
\ENSURE Optimal solution $x^*$ to~\eqref{eq:primal}.
\FORALL{$t=1, 2, \dots, T$} 
\STATE Pick an index set $\Lambda^t$ randomly without replacement, and compute
\[x^{t}=x^{t-1}- \frac{1}{t} \big[m^{-1}A^{\top}_{\Lambda^t}(A_{\Lambda^t}x^{t-1}-Y_{\Lambda^t}) + \lambda \cdot \sgn{x^{t-1}} \big]. \]
\ENDFOR
\STATE Set $x^* = x^T$.
\end{algorithmic}
\end{algorithm}

\vspace{0.05in}
\noindent{\bf Positive.} Compared to the full sub-gradient version, the memory cost is reduced to $O(md)$ where $m$ is the batch size that is much smaller than $n$. In particular, if we use the stochastic variant, memory footprint is only $O(d)$. Also, the mini-batch/stochastic variants are able to cope with streaming data since they update the solution dynamically. Finally, the computation in each iteration is cheap.

\vspace{0.05in}
\noindent{\bf Negative.}  One drawback is a worse convergence rate of $O(1/\sqrt{T})$, which is the price paid for the randomness~\cite{nesterov2004introductory}. Another specific drawback is that the gradient is noisy, i.e., large variance due to randomness, and hence a good $\lambda$ that works well for the batch method may not be suitable for the stochastic one since samples are different to each other. Finally, it turns out to be more difficult to expect sparse solutions due to the vanishing step size.

\subsection{Forward Backward Splitting~(Fobos)}
Fobos~\cite{fobos} is an interesting algorithm for solving the Lasso problem and tackles the shortcoming of sub-gradient methods. Recall that sub-gradient methods always compute the sub-gradient of $L_1$ norm and add it. The key idea of Fobos is updating the solution in two phases. First, it computes the gradient for the least-squares loss and performs gradient descent, i.e.,
  \begin{align}
  b^t = x^{t-1} - \eta_t (\ba_{i_t} \cdot x^{t-1} - y_{i_t}) \cdot \ba_{i_t}^{\top}.
  \end{align}
Note that the least-squares loss is differentiable. Hence, we move forward to a new point $b^t$. Second, it takes the $L_1$ norm into consideration by minimizing the following problem:
\begin{align}\label{eq:fobos_b} 
\min_x\quad \frac{1}{2} \twonorm{b^t - x} + \lambda \eta_t \onenorm{x},
\end{align}
where the closed-form solution is given by
\begin{align}
x^t = \mathcal{S}_{\lambda \eta_t}(b^t).
\end{align}

Intuitively,~\eqref{eq:fobos_b} is looking for a new iterate that interpolates between two goals: 1) stay close to the obtained point $b^t$ that minimizes the least-squares (in expectation), and 2) exhibit the sparsity pattern. As~\cite{fobos} suggested, the step size $\eta_t$ should be inversely proportional to $\sqrt{t}$, which is also the choice in Algorithm~\ref{alg:fobos}.

\begin{algorithm}[H]
\caption{Fobos}
\begin{algorithmic}[1]\label{alg:fobos}
\REQUIRE Training sample $(A, Y)$, parameter $\lambda$, maximum iteration count $T$.
\ENSURE Optimal solution $x^*$ to~\eqref{eq:primal}.
\FORALL{$t=1, 2, \dots, T$} 
\STATE Pick one sample $i_t$ randomly, and compute
\begin{align*}
b^t &= x^{t-1}- \frac{1}{\sqrt{t}} (\ba_{i_t} \cdot x^{t-1}- y_{i_t}) \ba_{i_t}^{\top},\\
x^{t} &= \mathcal{S}_{\lambda\eta_t} (b^t).
\end{align*}
\ENDFOR
\STATE Set $x^* = x^T$.
\end{algorithmic}
\end{algorithm}

\vspace{0.05in}
\noindent{\bf Positive.} First of all, it is guaranteed that the Fobos algorithm converges with the rate $O(1/\sqrt{T})$ as soon as the step size is $O(1/\sqrt{t})$. Second, since Fobos performs soft-thresholding, the solutions could be sparse. Finally, each step in Fobos can be computed efficiently and the memory cost is only $O(d)$.

\vspace{0.05in}
\noindent{\bf Negative.} Although Fobos appears promising, there are two issues regarding the sparsity: 1) Like the stochastic sub-gradient method, each time we only pick a random sample. Hence, it is not easy to find a large parameter $\lambda$ (the larger the more sparse) which works for all individual sample. 2) As the learning rate $\eta_t$ vanishes with $t$, it is evident that for the last iterations, the soft-thresholding step cannot contribute much to a sparse solution.

\vspace{0.1in}
\noindent{\bf Variants of Fobos.} To partially handle the first shortcoming above, we also implement a mini-batch version of Fobos. Since the implementation is easy, we omit details here. We also attempt a round based variant. That is, for the first $T/2$ iterations, we choose $0.5\lambda$, while for the last $T/2$ iteration we choose $\lambda$. This partially renders us some space to employing a larger $\lambda$.

\subsection{$L_0$ Constrained SGD}
While it is well known that the Lasso formulation~\eqref{eq:primal} is able to effectively encourage sparse solutions, it is actually impossible to characterize how sparse the Lasso solution is for a given $\lambda$. This makes tuning the parameter hard and time-consuming.

Alternatively, the $L_0$ norm counts the number of non-zero components and thereby, imposing an $L_0$ constraint guarantees sparsity. To this end, we are to solve the following non-convex program:
\begin{align}\label{eq:L0}
\min_{x \in \Rd}\quad G(x) \defeq \frac{1}{2n} \twonorm{Y - Ax}^2,\quad \st \zeronorm{x} \leq K,
\end{align}
where $K$ is a tunable parameter that controls the sparsity. Note that at this moment, we are only minimizing the least-squares loss, since the constraint already enforces sparsity. Again, the objective function $G(x)$ is decomposable:
\begin{align}
G(x) = \frac{1}{n} \sum_{i=1}^n g_i(x),
\end{align}
where
\begin{align}
g_i(x) \defeq \frac{1}{2} \twonorm{y_i - \ba_i \cdot x}^2.
\end{align}

Note that the problem above is non-convex and algorithms such as~\cite{blumensath2009iterative,needell2009cosamp} optimize it in a batch fashion. To be more detailed, these methods basically compute a full gradient followed by a hard thresholding step on the iterate. As mentioned earlier, it is not practical to evaluate a full gradient on a large dataset owing to computational and memory issues.

Henceforth, in this work, we attempt a stochastic gradient descent algorithm to solve~\eqref{eq:L0}. The high level idea follows both from the compressed sensing algorithms~\cite{blumensath2009iterative,needell2009cosamp} and the vanilla stochastic gradient method. That is, in each iteration $t$, given the previous iterate $x^{t-1}$, we randomly pick a sample $i_t$ to compute an unbiased stochastic gradient, followed by a usual gradient descent update:
\begin{align}
b^t = x^{t-1} - \eta_t \nabla g_{i_t}^{}(x^{t-1}).
\end{align}
Subsequently, to ensure the iterate satisfies the $L_0$ constraint, we perform hard thresholding $\Hk{\cdot}$ which sets all but the $K$ largest elements (in magnitude) to zero:
\begin{align}
x^t = \Hk{b^t}.
\end{align}

We summarize the algorithm in Algorithm~\ref{alg:L0}.
\begin{algorithm}[H]
\caption{$L_0$ constrained SGD}
\begin{algorithmic}[1]\label{alg:L0}
\REQUIRE Training sample $(A, Y)$, parameter $k$, maximum iteration count $T$.
\ENSURE Optimal solution $x^*$ to~\eqref{eq:L0}.
\STATE Initialize the iterate $x^0$ with zero vector.
\FORALL{$t=1, 2, \dots, T$}
\STATE Uniformly pick a sample $i_t$ and update
\[x^t = \Hk{x^{t-1} - \eta_t (\ba_{i_t} \cdot x - y_{i_t}) \ba_{i_t}^{\top}}.\]
\ENDFOR
\STATE Set $x^* = x^T$.
\end{algorithmic}
\end{algorithm}

\vspace{0.05in}
\noindent{\bf Positive.} The $L_0$ constrained SGD is the only method in this work that {\em ensures} a $k$-sparse solution. It is computationally efficient since the gradient descent step costs only $O(d)$ and the hard thresholding step costs $O(k\log d)$. It is also memory efficient, i.e., $O(d)$ memory suffices.

\vspace{0.05in}
\noindent{\bf Negative.} Perhaps the only shortcoming is the difficulty in characterizing the convergence rate, especially for realistic data where some standard assumptions like restricted strong convexity and restricted smoothness are not satisfied~\cite{agarwal2012fast}. However, our empirical study demonstrates that surprisingly, the non-convex formulation always converges fast.

To close this section, we give a comparison among all the methods in Table~\ref{tab:comp}.
\begin{table}[t]
\caption{{\bf Comparison of all algorithms.} Computation and memory cost for each iteration. Higher score for sparsity means a method tends to produce more sparse solutions.}
\centering
\resizebox{\linewidth}{!}{
\begin{tabular}{lcccc}
\toprule
Algorithm & Computation & Memory & Convergence & Sparsity\\
\midrule
CD & $O(nd)$ & $O(nd)$ & $1/T^2$ & $4$\\
SubGD & $O(nd)$ & $O(nd)$ & $1/T$ & $2$\\
Mini-SubGD & $O(md)$ & $O(md)$ & $1/\sqrt{T}$ & $2$\\
Stoc-SubGD & $O(d)$ & $O(d)$ & $1/\sqrt{T}$ & $1$\\
Fobos & $O(d)$ & $O(d)$ & $1/\sqrt{T}$ & $2$\\
Mini-Fobos & $O(md)$ & $O(md)$ & $1/\sqrt{T}$ & $3$\\
Round-Fobos & $O(d)$ & $O(d)$ & $1/\sqrt{T}$ & $2$\\
$L_0$-SGD & $O(d)$ & $O(d)$ & unknown & perfect\\
\bottomrule
\end{tabular}}\label{tab:comp}
\end{table}

\section{Experiments}
\subsection{Datasets}

To test our Lasso and $\ell_0$ implementations, we selected the Gisette dataset . The examples in this dataset are instances of the numeric characters $4$ and $9$ that must be classified into the appropriate numeric category. All digits were scale normalized and centered in a fixed image window of $28\times 28$ prior to feature extraction.

This dataset consists of $6000$ training examples, $3000$ positive and $3000$ negative. There are $1000$ testing examples, again $50\%$ positive and $50\%$ negative. Each example consists of $5000$ features. Critically though, only $2500$ of these features are real, the other $2500$ were added as dummy (random) features, meaning they have no predictive power. Therefore, any good Lasso solution for this dataset should set at least $50\%$ of the weights to $0$. We want to determine which algorithm (if any) will produce the best tradeoff between memory cost, accuracy and sparsity of the solution.

An important point regarding the $\ell_0$ constrained SGD is that the parameter $K$ essentially determines the sparsity of resulting solution since $K$ defines the number of weighting coefficients that will be set to $0$. For example, if we have $K=500$ in a dataset with $5000$ features, the sparsity will be $0.1$ since at most $10\%$ of the features can have non-zero weights.

\subsection{Setup}

Since Lasso requires tuning of the soft thresholding parameter $\lambda$, we tested our different Lasso implementations along a range of different $\lambda$ values (and in the case of the $\ell_0$ constrained SGD, the parameter $K$) until performance became asymptotic. $\lambda$ values ranged from $1.0\times 10^{-5}$ to $1.0\times 10^{2}$, while the $K$ parameter in the $\ell_0$ constrained SGD ranged from $20$ to $500$. For our mini-batch implementations, we used a batch size of $20$ random samples. We then recorded the accuracy and sparsity of each solution.  Accuracy is computed as classification accuracy (the percentage of test examples correctly classified as a $4$ or a $9$). Sparsity is given by the percentage of non-zero weight. For example, a sparsity value of $100$ would mean that none of the weights were set to $0$, likewise a sparsity value of $0.75$ would indicate that only $75\%$ of the weights are non-zero and $25\%$ are $0$. Therefore, lower values correspond to more sparse solution. Algorithms that provide solutions preferable in both accuracy and sparsity will be deemed better solutions.

\subsection{Results}
As can be seen in Table 2, all Lasso implementations were able to achieve reasonably high accuracy given proper parameter tuning. Maximal accuracy values ranged from $0.8810$ to $0.9700$ across the different Lasso implementations. However, only Coordinate Descent and the $\ell_0$ Constrained SGD were able to produce truly sparse solutions.

\begin{table}[h!]
\caption{Maximal accuracy and corresponding sparsity and $\lambda$/K value for each Algorithm.}
\centering
\begin{tabular}{l c c c}
\hline
Algorithm & Accuracy & Sparsity & $\lambda$/$K$ \\
\hline
Fobos & $0.9690$ & $0.9910$ & $1.0\times 10^{-5}$ \\
Fobos Round & $0.9700$ & $0.9910$ & $1.0\times 10^{-5}$ \\
Fobos Minibatch & $0.9680$ &  $0.9906$ & $1.0\times 10^{-5}$ \\
Coordinate Descent & $0.9350$ & $0.0738$ & $1.0\times 10^{-2}$ \\
Stochastic Subgradient & $0.8810$ & $0.9910$ & $1.0\times 10^{-5}$ \\
Minibatch Subgradient & $0.8860$ & $0.9910$ & $1.0\times 10^{-5}$ \\
L0 Constrained SGD & $0.9470$ & $0.0800$ & $K=400$ \\
\hline
\end{tabular}
\label{table: 1}
\end{table}

If we look at the sparsity of the solutions generated by each Lasso implementation in Figure 1 we notice a clear trend. As we should expect, as $\lambda$ increases, so does the sparsity. The one exception is the Subgradient Minibatch and Stochastic Subgradient (not plotted), which always result in dense solutions. Therefore, the Minibatch and Stochastic Subgradient Descent methods are not able to produce sparse solutions.

\begin{figure}[!ht]
\centering
\includegraphics[width=0.5\textwidth]{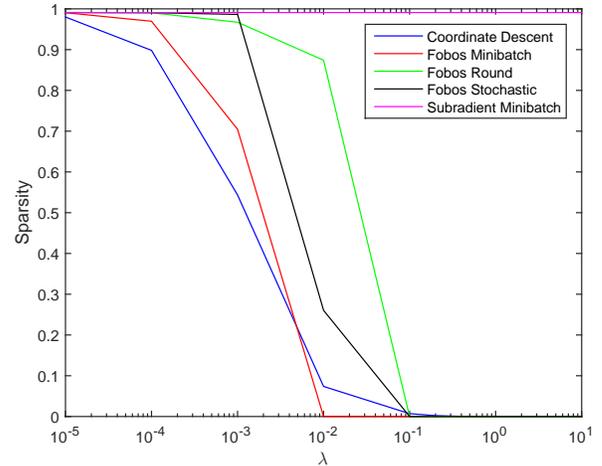}
\caption{Sparsity as a function of $\lambda$ for the majority of Lasso Implementations. Stochastic Subgradient Descent is not plotted since it lies on top of the Subgradient Minibatch results.}
\end{figure}

Although the various Fobos implementations are able to generate sparse solutions, the accuracy of these sparse solutions is extremely low (e.g. $50\%$). The only Lasso implementation plotted here that can produce both sparse and accurate solutions is Coordinate Descent, which has the significant drawback of being a batch method and therefore requiring extensive memory.

As previously mentioned, the $\ell_0$ constrained SGD will always produce a sparse solution since the sparsity is directly related to the parameter $K$. Since the sparsity of the solution for this implementation is always pre-defined, it is not necessary or informative to plot here.

\subsubsection{Batch Method}
Overall, our batch method (Coordinate Descent) performed quite well. We obtained sparse, accurate solutions for $\lambda$ ranging from $1.0\times 10^{-2}$ to $6.0\times 10^{-1}$. This is not surprising since the batch method computes the gradient for the entire dataset, but has the unfortunate drawback of being computationally and memory inefficient.
\subsubsection{Online Method}
Of the online methods, only the $\ell_0$ constrained SGD performed well. Stochastic Subgradient Descent was unable to produce sparse solutions for any setting of $\lambda$, while the Stochastic Fobos and Round Fobos implementations were unable to achieve sparse and accurate solutions. Instead, we obtained Fobos solutions that were either accurate or sparse but not both. $\ell_0$ constrained SGD was able to produce solutions that were both sparse and accurate for $K$ ranging from $20$ (corresponding to sparsity of $4.0\times 10^{-3}$) to $500$ (corresponding to a sparsity of $1.0\times 10^{-1}$).

For the Fobos Round algorithm (Figure 2), the best convergence was found for small values of $\lambda$. Additionally, the value of the objective function was highly variable from iteration to iteration, and therefore the error does not decrease monotonically. Similar effects were found for the stochastic variant of Fobos (Figure 3), with similar rates of convergence.

\begin{figure}[!ht]
\centering
\includegraphics[width=0.5\textwidth]{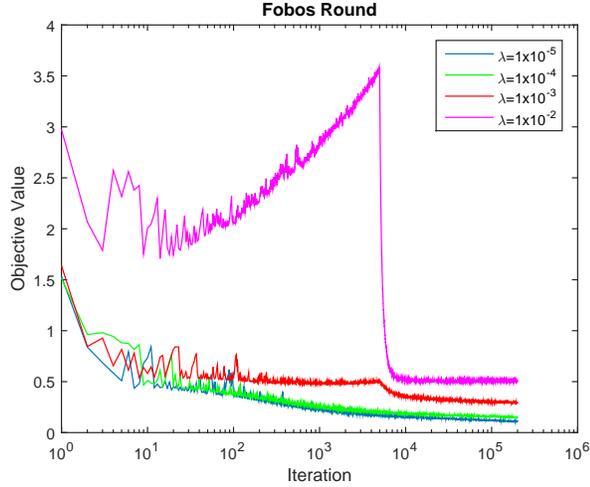}
\caption{Convergence rate for the Fobos Round algorithm.}
\end{figure}

\begin{figure}[!ht]
\centering
\includegraphics[width=0.5\textwidth]{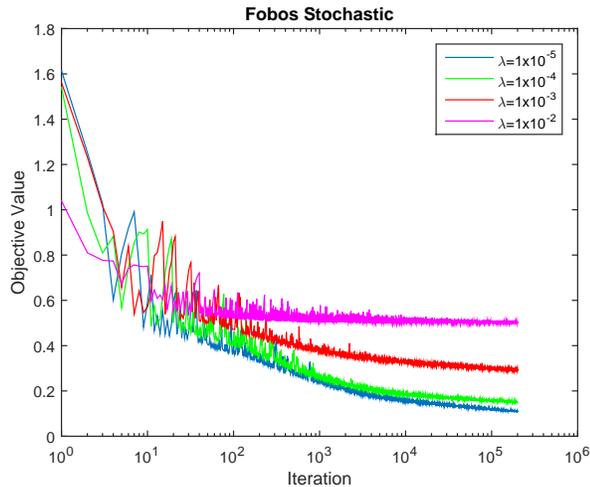}
\caption{Convergence rate for the Fobos Stochastic algorithm.}
\end{figure}

Similar to the online implementations of Fobos, the stochastic Subgradient Descent (Figure 4) had the best convergence rates for small $\lambda$. On the other hand, the rate of convergence for the stochastic SubGD is much slower than for Fobos, so even though the value of the objective function does decrease monotonically, the stochastic Subgradient Descent provides a worse solution than the online implementations of Fobos.

\begin{figure}[!ht]
\centering
\includegraphics[width=0.5\textwidth]{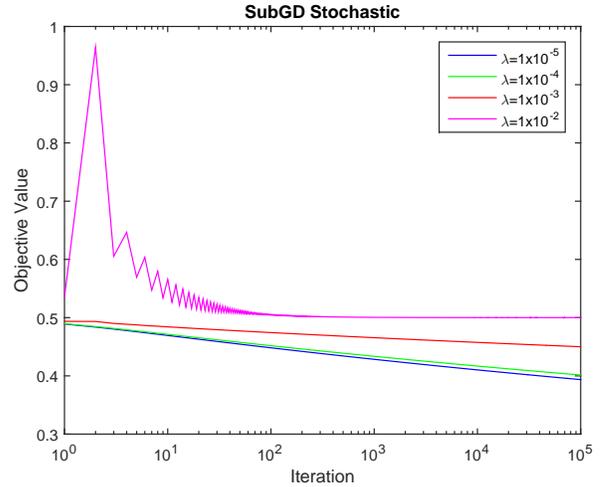}
\caption{Convergence rate for the Stochastic Subgradient Descent algorithm.}
\end{figure}

The $\ell_0$ constrained SGD (Figure 5) shows the best rate of convergence out of all the online algorithms. We found the best convergence rate for larger values of $K$, although there is probably a point of diminishing returns where $K$ becomes too large and we get a worse solution. Similar to the implementations of Fobos, the value of the objective function for $\ell_0$ constrained SGD is highly variable from iteration to iteration due to variacne.

\begin{figure}[!ht]
\centering
\includegraphics[width=0.5\textwidth]{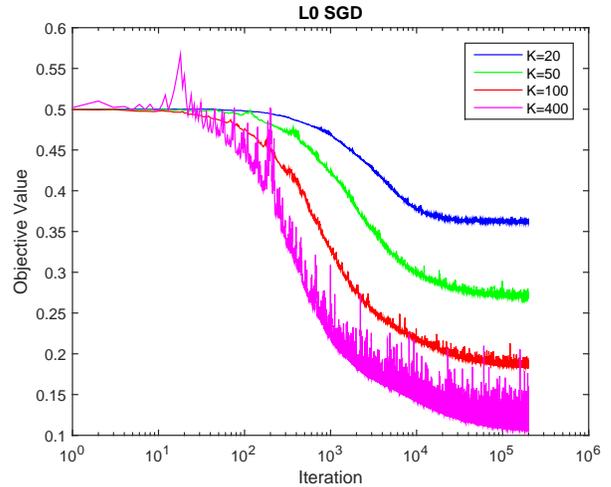}
\caption{Convergence rate for the $\ell_0$ constrained SGD algorithm.}
\end{figure}

\subsubsection{Mini-batch Method}
Both mini-batch methods were unable to produce good solutions. Similar to the stochastic version, the mini-batch version of Subgradient Descent was also unable to produce any sparse solutions. Additionally, similar to the other Fobos implementations, the mini-batch version was able to produce solutions that were either sparse or accurate, but not both.

For the two mini-batch methods (Fobos and Subgradient Descent) we see similar effects for the rate of convergence. Once again, small $\lambda$ corresponds to faster convergence. Additionally, the value of the objective function decreases monotonically for both mini-batch algorithms. However, as in the online implementations, the mini-batch version of Fobos (Figure 6) converges at a much faster rate than the mini-batch version of Subgradient Descent (Figure 7).

\begin{figure}[!ht]
\centering
\includegraphics[width=0.5\textwidth]{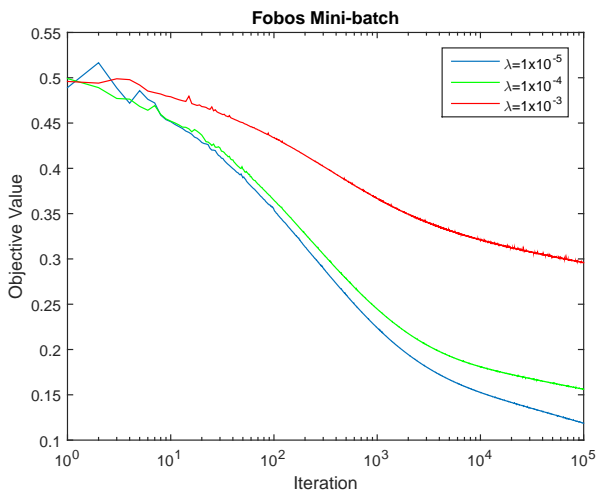}
\caption{Convergence rate for the Fobos Mini-batch algorithm.}
\end{figure}

\begin{figure}[!ht]
\centering
\includegraphics[width=0.5\textwidth]{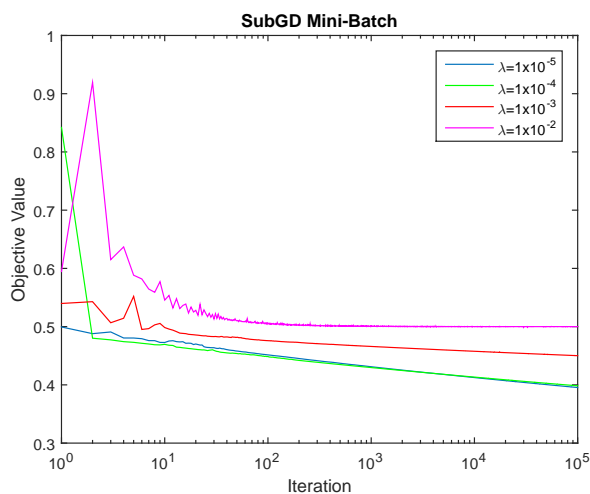}
\caption{Convergence rate for the Mini-batch Subgradient Descent algorithm.}
\end{figure}

\section{Conclusion and Future Works}
We investigated the quality of solutions, measured in terms of accuracy and sparsity, that are produced by various batch, mini-batch, and online implementations of Lasso and a non-convex $\ell_0$ constrained SGD. Unsurprisingly, we found that the batch method implemented here, Coordinate Descent, was able to produce solutions that were both sparse and accurate. Although Coordinate Descent produces good solutions, it is a batch method and therefore has the side effect of requiring a large amount of memory to compute the solution. Additionally, the cost of computing the gradient on each iteration is much larger than for the online algorithms.

Mini-batch and stochastic methods have the important advantage of requiring very little memory to compute solutions. As previous work has shown, the drawback is that the solutions produced are usually only either accurate or sparse, but not both. In the case of Subgradient Descent, the mini-batch and stochastic versions of the algorithm were not able to produce sparse solutions. Additionally, we were not able to obtain solutions that were both sparse and accurate for any of the Fobos implementations. However, the $\ell_0$ constrained SGD algorithm performed quite well, producing solutions that were both accurate and sparse. Perhaps somewhat surprisingly, we found that the $\ell_0$ constrained SGD algorithm also had the fastest rate of convergence among the online methods. By using hard thresholding (e.g. setting all but the $K$ largest absolute values to $0$) we were able to develop an online algorithm that guarantees we will obtain sparse solutions.

Although the $\ell_0$ constrained SGD algorithm is guaranteed to produce solutions that are sparse, it remains to be seen whether it will produce accurate solutions on different data sets. As mentioned previously, $50\%$ of the features in the Gisette data set used here were dummy features, and were therefore uninformative for classifying each example. Because of this, the Gisette data set is very well-suited for using Lasso for variable selection since the structure of the data set naturally promotes sparse solutions. It is necessary to test additional data sets to ensure the generalizability of our results.

One important advantage of the $\ell_0$ constrained implementation of SGD is the intuitive setting of the $K$ parameter as compared with $\lambda$ in the other algorithms. This is because $K$ corresponds directly to the sparsity of the solution, so setting the parameter $K$ is really just defining the sparsity of the solution. This becomes critical in settings where we have prior knowledge about how sparse our solution should be. Rather than testing a multitude of different $\lambda$, one could instead test a much smaller set of $K$, allowing us to obtain a solution more efficiently.

\bibliographystyle{abbrv}

\end{document}